\documentclass[letterpaper]{article} 
\usepackage{aaai2027}  
\usepackage[hyphens]{url}  
\usepackage{graphicx} 
\usepackage{natbib}  
\usepackage{caption} 
\usepackage{algorithm}
\usepackage{algorithmic}

\usepackage{newfloat}
\usepackage{listings}
\usepackage{times}
\usepackage{helvet}
\usepackage{courier}
\usepackage{microtype}
\usepackage{multirow}
\usepackage{amsmath}
\usepackage{amssymb}
\usepackage{graphicx}
\usepackage{xcolor}
\usepackage{tikz}
\usepackage{pgfplots}
\DeclareCaptionStyle{ruled}{labelfont=normalfont,labelsep=colon,strut=off} 
\floatstyle{ruled}
\newfloat{listing}{tb}{lst}{}
\floatname{listing}{Listing}

\usepackage{booktabs}

\title{GRAFT: Adaptive DLM-Based Draft Tree Construction with \\Target-Distilled Edge Scoring}
\author{
    Xuming Ye\textsuperscript{\rm 1}, Zeming Ma\textsuperscript{\rm 1}, Runjie Yu\textsuperscript{\rm 1}, Yuan Liu\textsuperscript{\rm 1,3}, Tianle Li\textsuperscript{\rm 2}\\
    Shuhan Bai\textsuperscript{\rm 1}, Jian Zhou\textsuperscript{\rm 1}\corresponding, Fei Wu\textsuperscript{\rm 1}\corresponding
}
\affiliations{
    \textsuperscript{\rm 1}HuaZhong University of Science and Technology, Wuhan, China\\
    \textsuperscript{\rm 2}The University of Hong Kong, Hong Kong, China\\
    \textsuperscript{\rm 3}CECloud Computing Technology Co., Ltd, Wuhan, China\\

    xumingye@hust.edu.cn, mzm@hust.edu.cn, d202381500@hust.edu.cn, liu\_yuan@hust.edu.cn, tianleli@connect.hku.hk, \\shuhanbai0329@hust.edu.cn, jianzhou@hust.edu.cn, wufei@mail.hust.edu.cn
}

\newcommand{\ddtree}{DDTree}
\newcommand{\dflash}{DFlash}
\newcommand{\tdes}{TDES}
\newcommand{\saba}{SABA}

\newcommand{\graft}{GRAFT}
\newcommand{\dlm}{DLM}

\begin{document}

\maketitle

\begin{abstract}

Tree-based speculative decoding raises the mean accepted tokens of standard speculative decoding by verifying multiple draft paths, and existing tree builders typically construct these paths through parent-conditioned expansion, where each child token is generated conditioned on its parent path. This construction is incompatible with diffusion language model (\dlm{}) drafters such as \dflash{}, which produces all future-position distributions in a single forward pass. \ddtree{} bridges this gap by treating high-probability tokens from each future-position distribution as candidate nodes and selecting edges between consecutive positions under a fixed node budget. However, its edge selection relies on token probability alone without modeling parent--child compatibility, so target-compatible tokens can be attached to wrong parents; moreover, its fixed budget ignores that the throughput-optimal tree size varies with the decoding state. We propose \graft{}, a draft-tree construction framework for \dlm{}-based speculative decoding. \graft{} introduces Target-Distilled Edge Scoring (\tdes{}), which distills parent--child preferences from target-model traces to select target-compatible edges, and State-Aware Budget Allocation (\saba{}), which sets the per-round tree budget by balancing expected draft gain against verification cost. Across multiple models and tasks, \graft{} achieves $2.13\times$--$6.36\times$ end-to-end speedup over autoregressive decoding while adding less than $0.5$\,ms of overhead per round, approximately $1.4\%$ of the target-model verification latency.


\end{abstract}


\section{Introduction}

Large language models (LLMs) have become core infrastructure across applications ranging from dialogue and code generation to agentic systems \cite{openai2026gpt5,DBLP:journals/corr/abs-2501-12948,DBLP:conf/iclr/HuLC25}. Their inference efficiency, however, is fundamentally limited by autoregressive decoding, which outputs a single token per forward pass, leaving much of the GPU's parallel capacity underutilized. Speculative decoding relieves this bottleneck through a draft-then-verify paradigm: a lightweight draft model proposes a candidate sequence as a draft, and the target model verifies the draft in parallel, committing the longest prefix consistent with the target model \cite{DBLP:journals/corr/abs-2302-01318,DBLP:conf/icml/LeviathanKM23}. The resulting speedup of speculative decoding is largely driven by the mean accepted tokens (MAT).



Tree-based speculative decoding further improves MAT by replacing a single draft sequence with a draft tree \cite{DBLP:conf/asplos/MiaoOZCWZWZYSSC24,DBLP:conf/icml/CaiLGPLCD24,li2026eagle}. With a tree attention mask \citep{DBLP:conf/asplos/MiaoOZCWZWZYSSC24}, the target model can verify the draft tree in a single forward pass. In most model-based systems, the tree is constructed through autoregressive \textit{parent-conditioned} node expansion, where the drafter widens or deepens the tree by generating each child token under the path of its parent \cite{li2026eagle,DBLP:journals/tacl/WangSLXYDWZ25,DBLP:journals/corr/abs-2510-26577,DBLP:journals/corr/abs-2601-07353}.


However, diffusion language model (DLM) drafters do not fit this parent-conditioned construction. For instance, the state-of-the-art drafter, DFlash, predicts future-token distributions in a single forward pass \citep{DBLP:journals/corr/abs-2602-06036}. This forward pass is conditioned on the current target-model states, so expanding a new parent path would require target-model states that are not available before verification. Therefore, the problem becomes how to assemble a draft tree from one-pass outputs without re-running the drafter for each parent path.




To construct a draft tree from \dflash{}, \ddtree{} adopts a \textit{parent-agnostic} tree assembly strategy. It first treats high-probability tokens at each future position as candidate nodes. It then connects nodes across consecutive positions to form candidate paths. Each path is scored only by multiplying the corresponding per-position token probabilities, and \ddtree{} selects high-scoring paths under a \textit{fixed} node budget. This assembly is parent-agnostic since a child token contributes the same score regardless of which parent path it extends. We claim that the parent-agnostic and fixed-budget design introduces two challenges: (i) \textit{parent--child mismatch}, and (ii) \textit{budget misallocation}.



\begin{figure*}[t]
  \centering
  \includegraphics[width=\textwidth]{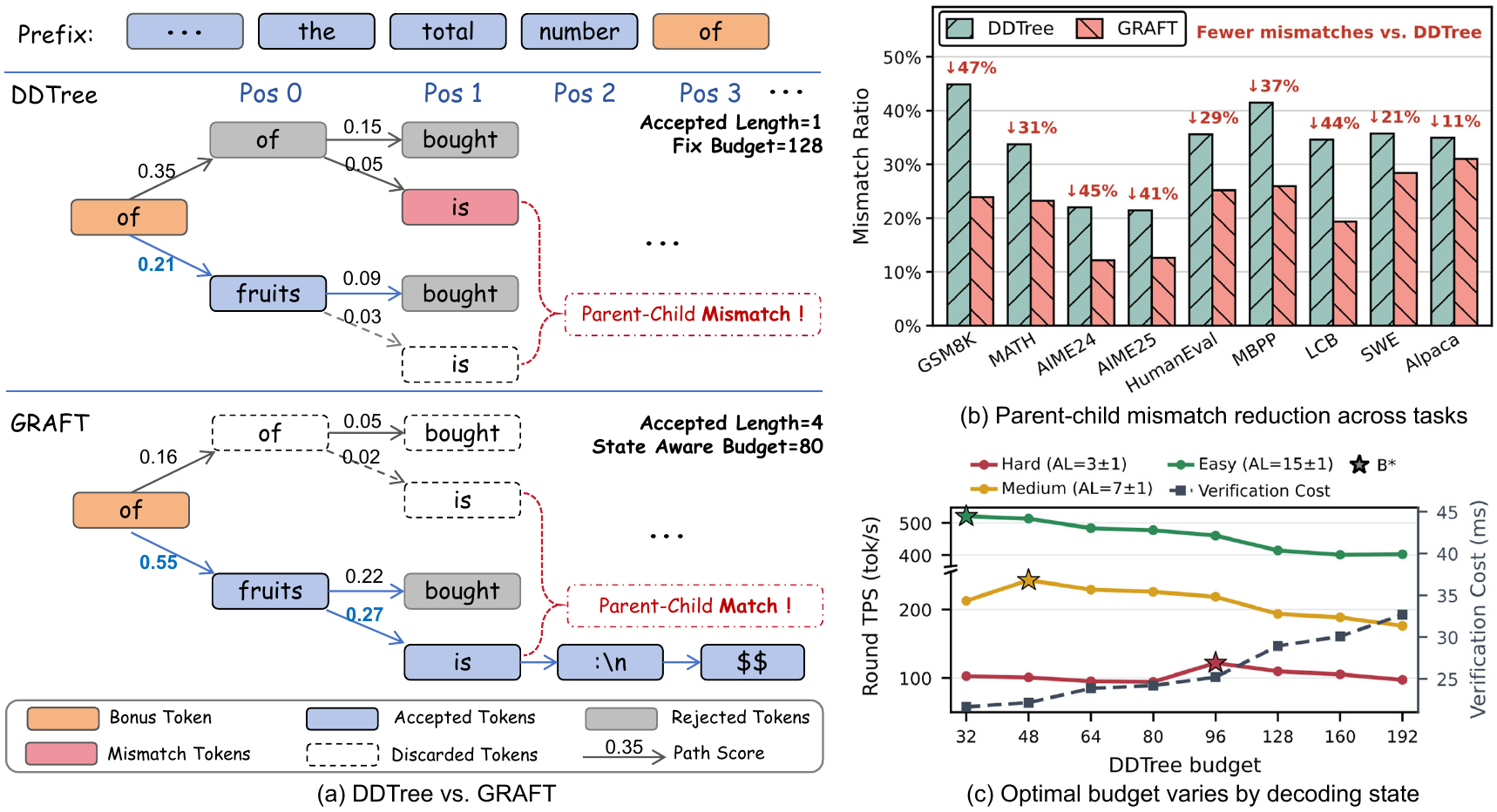}

  \caption{Architectural comparison and empirical analysis of DLM-based tree construction. (a) \ddtree{} assembles draft trees from position-wise \dflash{} marginals, so a target-compatible child can be selected under the wrong parent. Here \ddtree{} keeps the edge \texttt{of}$\rightarrow$\texttt{is} and discards the target-compatible edge \texttt{fruits}$\rightarrow$\texttt{is}, whereas \graft{} prefer target-compatible edges and accepts three more tokens than \ddtree{}. (b) Parent--child mismatch occurs across various tasks, and \graft{} consistently reduces the mismatch rate over \ddtree{} (Qwen3-8B). (c) The throughput-optimal tree budget $B^*$ depends on the decoding state.}
    
  
  \label{fig:motivation}
\end{figure*}

\noindent \textbf{Parent--Child Mismatch.} The path score reflects only the draft model's confidence in each token in isolation, not how compatible a child is with a specific parent. It turns out that a target-compatible token is present in the tree but attached to the wrong parent, which we call \textit{parent--child mismatch}. As illustrated in Figure~\ref{fig:motivation}(a), DDTree keeps the high-scoring path \texttt{of}$\rightarrow$\texttt{is}, and discards the target-compatible edge \texttt{fruits}$\rightarrow$\texttt{is}, since \texttt{is} contributes the same score under both parents and the path through \texttt{of} has the higher score. Figure~\ref{fig:motivation}(b) further shows that parent--child mismatch appears across tasks, indicating that \dlm{}-based draft-tree construction must explicitly model parent--child compatibility.


\noindent \textbf{Budget Misallocation.} \ddtree{} assigns the same node budget to every decoding round, while the value of additional nodes varies with the current decoding state. Easy rounds already obtain long accepted paths, so extra nodes mainly add verification cost; hard rounds start with shorter accepted paths, where a larger tree can expose useful candidates but only until the accepted-length gain saturates. Figure~\ref{fig:motivation}(c) shows this state-dependent trade-off. The throughput-optimal budget differs across easy, medium, and hard rounds, whereas verification cost keeps increasing with tree size. Thus, a fixed budget misallocates verification work across rounds, motivating a per-round budget decision.


To that end, we propose \textbf{\graft{}}, a tree construction framework for \dlm{}-based speculative decoding comprising Target-Distilled Edge Scoring (\tdes{}) and State-Aware Budget Allocation (\saba{}). \graft{} builds on two observations: target-model traces reveal preferences over parent--child transitions, and the throughput-optimal tree budget varies with the current decoding state. Based on the first observation, \tdes{} distills a lightweight edge scorer from target model traces to select target-compatible edges, reducing parent--child mismatch across tasks. Based on the second observation, \saba{} models the trade-off between draft gain and verification cost, selecting the optimal budget that maximizes decoding throughput. In summary, \graft{} determines which edges enter the tree and how large the tree should be, as illustrated in Figure~\ref{fig:motivation}(a). Our main contributions are as follows:

\begin{itemize}
\item We identify two main challenges in \dlm{}-based draft-tree construction, namely parent--child mismatch and budget misallocation, and show that they respectively stem from parent-agnostic marginal assembly and fixed-budget of tree construction.
\item We propose \graft{}, a tree construction framework that combines Target-Distilled Edge Scoring for parent-conditional edge selection and State-Aware Budget Allocation for per-round tree sizing.
\item We conduct experiments across various models and datasets. The results show that \graft{} consistently improves end-to-end inference throughput by 2.13--6.36$\times$, while adding less than 0.5\,ms of overhead per round.
\end{itemize}

\section{Preliminaries}
\label{sec:prelim}

\paragraph{Speculative decoding.} Let $x^{(t)}$ denote the prefix at the beginning of decoding round $t$. Standard speculative decoding uses a lightweight draft model to propose a length-$K$ candidate sequence $\hat y_{t,1:K}$ and then uses the target model to verify the sequence in parallel \citep{DBLP:journals/corr/abs-2302-01318,DBLP:conf/icml/LeviathanKM23}. In greedy decoding notation, the candidate sequence is generated autoregressively as
\begin{equation}
  \hat y_{t,d}\sim P_{\rm draft}(\cdot\mid x^{(t)},\hat y_{t,<d}),\qquad d=1,\dots,K.
  \label{eq:standard-draft}
\end{equation}
We define $a_t$ as the number of tokens accepted at round $t$ and the mean acceptance tokens (MAT) as $\frac{1}{T}\sum_{t=1}^{T}a_t$ over $T$ rounds.

\begin{figure}[t]
  \centering
  \includegraphics[width=\columnwidth]{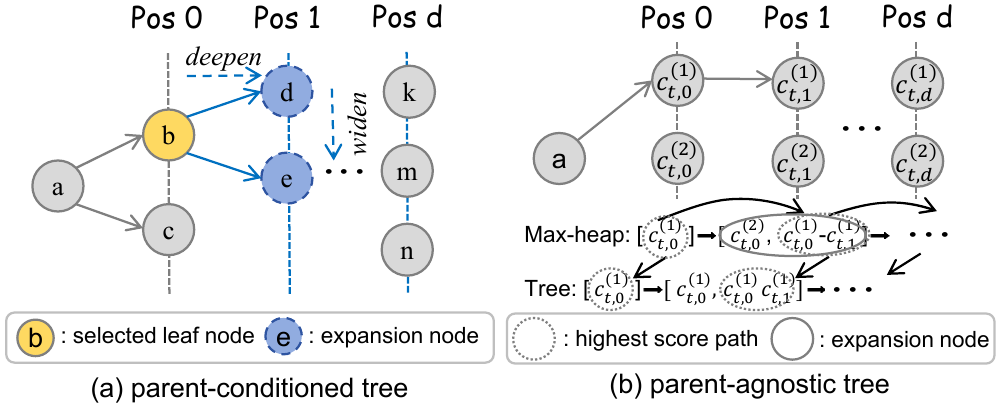}
  \caption{Parent-conditioned and parent-agnostic draft-tree construction.} 
  \label{fig:preliminary-tree-construction}
\end{figure}

\paragraph{Tree-based speculative decoding.} Tree-based speculative decoding aims to improve MAT by replacing a single draft sequence with a prefix-closed draft tree $\mathcal{T}_t$. For autoregressive drafters, $\mathcal{T}_t$ is commonly constructed through parent-conditioned node expansion, where each child token is generated conditioned on its parent path. As shown in Figure~\ref{fig:preliminary-tree-construction}(a), the tree builder can widen the tree by adding alternative children to a node or deepen it by recursively expanding a selected leaf node. The target model verifies all nodes in one forward pass using a tree-attention mask \citep{DBLP:conf/asplos/MiaoOZCWZWZYSSC24} and commits the longest accepted draft path.



\paragraph{\dflash{} one-pass marginals.} \dflash{} is a diffusion language model drafter that produces an entire draft block in one forward pass \citep{DBLP:journals/corr/abs-2602-06036}. Given $x^{(t)}$ and the current target-model states, \dflash{} outputs a token distribution $q_{t,d}(\cdot)$ over $\mathcal{V}$ for each depth $d=1,\dots,D$. Thus, $q_{t,d}(c)$ is the score of token $c$ at depth $d$, i.e., the token probability of $c$. Let $c_{t,d}^{(r)}$ denote the token with the $r$-th highest score at depth $d$, and \dflash{} selects $c_{t,d}^{(1)}$ at each depth as the draft,
\begin{equation}
  \hat y_{t,d}=c_{t,d}^{(1)}=\arg\max_{c\in\mathcal{V}}q_{t,d}(c),\qquad d=1,\dots,D.
  \label{eq:dflash-top1}
\end{equation}

\paragraph{\ddtree{} parent-agnostic tree construction.} Given the \dflash{} outputs, \ddtree{} constructs a prefix-closed draft tree under a budget $B$. For each candidate prefix $u=(u_1,\dots,u_\ell)$, it defines the path score as \citep{DBLP:journals/corr/abs-2604-12989}
\begin{equation}
  S_{\rm DD}(u)=\sum_{d=1}^{\ell}\log q_{t,d}(u_d).
  \label{eq:ddtree-path-score}
\end{equation}
As shown in Figure~\ref{fig:preliminary-tree-construction}(b), \ddtree{} maintains candidate prefixes in a max-heap and repeatedly expands the highest-scoring prefix to the tree until $B$ nodes have been selected. Appending token $c$ at depth $d$ always contributes $\log q_{t,d}(c)$ to the path score, regardless of its parent path. \ddtree{} is therefore parent-agnostic when scoring edges.


\section{Method}

\subsection{Overview}

\begin{figure*}[t]
  \centering
  \includegraphics[width=0.8\textwidth]{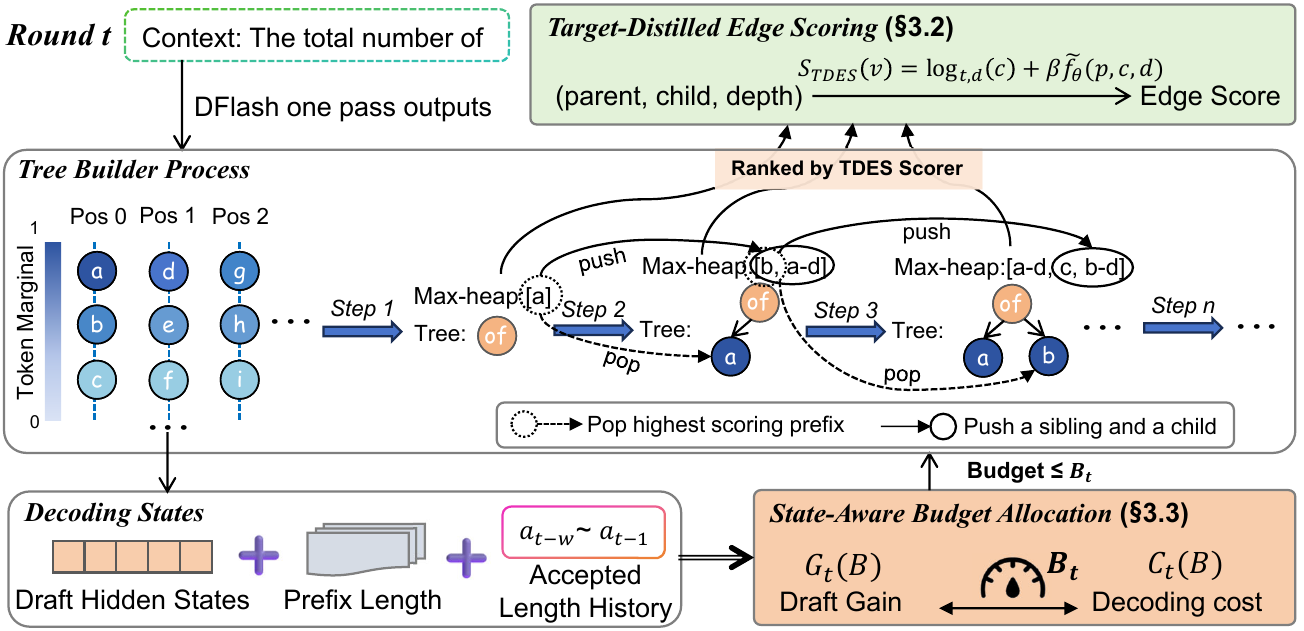}
  \caption{\textbf{\graft{} overview.} Given the one-pass outputs of \dflash{} and the current decoding state, \tdes{} and \saba{} produce parent-conditional edge scores $S_{\mathrm{TDES}}(v)$ and a per-round tree budget $B_t$, respectively. The \graft{} Tree Builder uses the edge scores to rank candidate prefixes and $B_t$ to terminate heap expansion before passing the resulting draft tree to the target model.}
  \label{fig:overview}
\end{figure*}

To address the challenges of parent--child mismatch and budget misallocation in \ddtree{}, we propose \graft{}, a tree construction framework for \dlm{}-based speculative decoding. Figure~\ref{fig:overview} illustrates the pipeline of \graft{}. Given the one-pass outputs of \dflash{} and the current decoding state, Target-Distilled Edge Scoring (\tdes{}) assigns parent-conditional scores to candidate edges, while State-Aware Budget Allocation (\saba{}) determines the tree budget $B_t$. Following \ddtree{}, the tree builder maintains candidate prefixes in a max-heap, ranks them using the \tdes{} scores, and stops expanding the tree when $B_t$ nodes have been selected.

\subsection{Target-Distilled Edge Scoring}
\label{sec:tdes}

To address the parent-agnostic edge scoring, \tdes{} keeps the token probability as the base score and adds a residual compatibility term for each parent--child edge. For an edge $v=(p,c,d)$ from parent token $p$ to child token $c$ at depth $d$, let $\mathcal{C}_{p,d}$ denote the children considered under $p$. We define
\begin{align}
S_{\mathrm{TDES}}(v)
  &= \log q_{t,d}(c)+\beta\,\widetilde f_\theta(p,c,d), \label{eq:tdes-score}\\
\widetilde f_\theta(p,c,d)
  &= f_\theta(p,c,d)-\frac{1}{|\mathcal{C}_{p,d}|}\sum_{c'\in\mathcal{C}_{p,d}}f_\theta(p,c',d), \label{eq:tdes-center}
\end{align}
where $f_\theta(p,c,d)$ estimates parent--child compatibility, and $\beta$ controls the strength of the residual correction. The centered score $\widetilde f_\theta$ subtracts the mean compatibility among candidate children under the same parent, removing parent-specific offsets while preserving relative deviations within $\mathcal{C}_{p,d}$. 

\begin{figure}[t]
  \centering
  \includegraphics[width=\columnwidth]{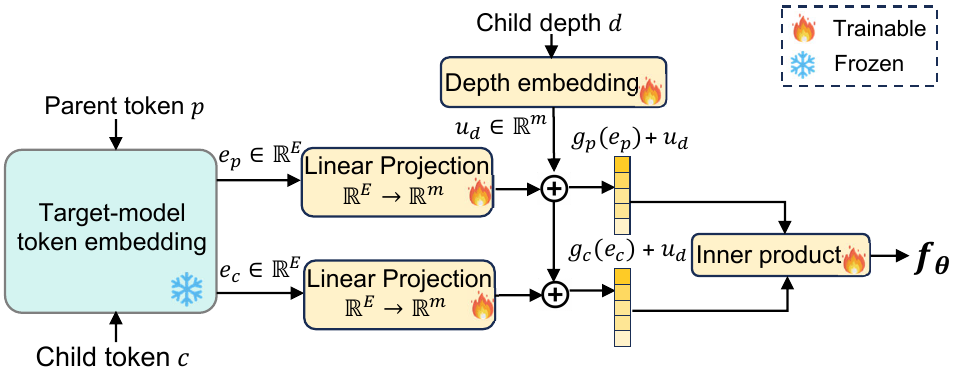}
  \caption{TDES scorer architecture. Trainable denotes parameters updated during \tdes{} training, while Frozen denotes parameters kept fixed.}
  \label{fig:scorer}
\end{figure}

\paragraph{Scorer architecture.} The edge scorer is invoked many times during heap expansion, so its runtime cost must remain negligible compared with a drafter or verifier forward pass. We therefore instantiate $f_\theta$ as a lightweight and cacheable bi-tower model over the immediate parent token, child token, and depth, as illustrated in Figure~\ref{fig:scorer}. 

Let $e_p,e_c\in\mathbb{R}^{E}$ be the frozen token embeddings of the parent and child from the target model. We map them into an $m$-dimensional scoring space with two linear projections $g_p,g_c:\mathbb{R}^{E}\rightarrow\mathbb{R}^{m}$. We add a depth embedding $u_d\in\mathbb{R}^{m}$ and score the edge by
\begin{equation}
f_\theta(p,c,d)=s\,\frac{\big(g_p(e_p)+u_d\big)^\top\big(g_c(e_c)+u_d\big)}{\sqrt{m}},
\label{eq:tdes-bitower}
\end{equation}
where $s>0$ is a learnable scale parameter. Since the projected token embeddings $g_p(e)$ and $g_c(e)$ depend only on the token id, we compute them once and reuse them during decoding.

\paragraph{Training.} We train $f_\theta$ from target-model traces collected during speculative decoding with \ddtree{}. The training objectives are as follows:

\emph{Distribution level.} For each accepted parent $p$ at depth $d-1$, let $z^{\mathrm{tgt}}_{p,c}$ be the target-model logit for child $c\in\mathcal{C}_{p,d}$. We define the target preference distribution and the corrected draft distribution as
\begin{align}
  \pi^{\mathrm{tgt}}_{p,d}(c)
  &=\underset{c\in\mathcal{C}_{p,d}}{\mathrm{softmax}}\!\left(z^{\mathrm{tgt}}_{p,c}\right),\\
  \pi^\theta_{p,d}(c)
  &=\underset{c\in\mathcal{C}_{p,d}}{\mathrm{softmax}}\!\left(\log q_{t,d}(c)+f_\theta(p,c,d)\right).
\end{align}
and align them with
\begin{equation}
  \mathcal{L}_{\mathrm{KL}}=\mathbb{E}_{(p,d)}\!\left[\mathrm{KL}\!\left(\pi^{\mathrm{tgt}}_{p,d}\,\Vert\,\pi^\theta_{p,d}\right)\right].
\end{equation}

\emph{Pairwise level.} For each mismatch trace, we form a target-compatible edge $v^+$ and an incompatible alternative $v^-$. The negative edge either attaches the same child to a wrong parent or a competing child to the correct parent. We enforce $f_\theta(v^+)>f_\theta(v^-)$ with a margin loss:
\begin{equation}
\mathcal{L}_{\mathrm{pair}}=\mathbb{E}_{(v^+,v^-)}\!\left[\operatorname{softplus}\!\left(\gamma-f_\theta(v^+)+f_\theta(v^-)\right)\right].
\end{equation}

\emph{Edge level.} For every verified edge $v=(p,c,d)$, we set $y_v=1$ when $c$ is the target next token under parent $p$ and $y_v=0$ otherwise. We use a class-balanced binary cross-entropy, where $w_1$ offsets the sparsity of positive edges:
\begin{equation}
\begin{aligned}
\mathcal{L}_{\mathrm{BCE}}=-\mathbb{E}_{(v,y_v)}\!\big[&w_1y_v\log\sigma(f_\theta(v))\\
&+(1-y_v)\log\!\left(1-\sigma(f_\theta(v))\right)\big].
\end{aligned}
\end{equation}

The final objective is
\begin{equation}
\mathcal{L}=\lambda_{\mathrm{KL}}\mathcal{L}_{\mathrm{KL}}+\lambda_{\mathrm{pair}}\mathcal{L}_{\mathrm{pair}}+\lambda_{\mathrm{BCE}}\mathcal{L}_{\mathrm{BCE}},
\label{eq:tdes-training}
\end{equation}
where $\lambda_{\mathrm{KL}}$, $\lambda_{\mathrm{pair}}$, and $\lambda_{\mathrm{BCE}}$ control the relative contributions of the three supervision signals.

\paragraph{Inference.} At inference time, \tdes{} assigns each candidate edge the score in Equation~\eqref{eq:tdes-score}. For a prefix $u$ containing edges $v_1,\ldots,v_{|u|}$, the path score is
\begin{equation}
S_{\mathrm{TDES}}(u)=\sum_{d=1}^{|u|}S_{\mathrm{TDES}}(v_d).
\label{eq:tdes-path-score}
\end{equation}
As shown in Figure~\ref{fig:overview}, the tree builder uses $S_{\mathrm{TDES}}(u)$ to rank the candidate prefixes in a max-heap and repeatedly adds the highest-scoring prefix to the tree until reaching the budget. In our tests, \tdes{} adds only $0.3$\,ms per decoding round, approximately 1.3\% of the target model verification latency.


\subsection{State-Aware Budget Allocation}
\label{sec:saba}

The optimal tree budget varies across decoding rounds as additional nodes offer different throughput utility under different decoding states. A larger tree can cover target-compatible sequences that would otherwise be missed, yet further expansion increases per-round decoding cost. To capture this trade-off, \saba{} formulates budget allocation as a state-conditioned optimization problem. 

\begin{equation}
  B_t=\arg\max\ U_t(B),\qquad U_t(B)=\frac{G_t(B)}{C_t(B)}.
  \label{eq:saba-objective}
\end{equation}

\paragraph{Problem formulation.} Given the current decoding state, \saba{} models the expected draft gain $G_t(B)$ and per-round decoding cost $C_t(B)$ as functions of the tree budget. Their ratio defines the throughput utility $U_t(B)$. \saba{} aims to find the optimal tree budget $B_t$ that maximizes $U_t(B)$ for the current decoding round.


\paragraph{Decoding state.} Following previous adaptive tree construction \citep{DBLP:journals/tacl/WangSLXYDWZ25,ning2026cas}, draft confidence and acceptance history serve as standard proxies for draft quality. We also include a \dflash{}-specific signal, \textit{representation degeneration}. In \dflash{}, adjacent hidden states tend to become more similar toward later positions, which makes later draft tokens less distinguishable. Thus, \saba{} characterizes each round using three complementary signals: draft confidence, representation degeneration, and acceptance history.

Let $h_{t,d}$ be the \dflash{} hidden state at depth $d$, and $r_{t,d}=\cos(h_{t,d},h_{t,d+1})$ for $d=1,\dots,D-1$. We define

\begin{equation}
\begin{gathered}
\mathrm{conf}_t
  =\frac{1}{D}\sum_{d=1}^{D}q_{t,d}\!\left(c_{t,d}^{(1)}\right),\quad
\mathrm{hist}_t
  =\frac{1}{w}\sum_{i=1}^{w}a_{t-i},\\
\mathrm{degen}_t
  =\frac{1}{D-1}\sum_{d=1}^{D-1}r_{t,d}
    +(r_{t,D-1}-r_{t,1}).
\end{gathered}
\end{equation}
where the two terms in $\mathrm{degen}_t$ capture the average adjacent-position similarity and its increase from early to late positions, and $w$ is the acceptance history window size. Together they form the decoding state $s_t=[\mathrm{conf}_t,\mathrm{degen}_t,\mathrm{hist}_t]$.

\paragraph{Gain model.} Draft gain increases with the tree budget but exhibits diminishing marginal returns. \saba{} therefore models its dependence on the tree budget $B$ and the current decoding state $s_t$ using a state-conditioned saturation curve.
\begin{equation}
G_t(B)
=1-\exp\!\left(-\frac{B}{\boldsymbol{\phi}^{\top}s_t}\right),
\label{eq:saba-gain}
\end{equation}
where $\boldsymbol{\phi}$ is a parameter set. $\boldsymbol{\phi}^{\top}s_t$ determines how quickly the gain saturates under the current decoding state. A larger value indicates that additional nodes may remain useful over a wider budget range, whereas a smaller value indicates earlier diminishing returns.


\paragraph{Decoding cost.} The per-round decoding cost consists of tree construction and target-model verification. For a fixed prefix length, both costs grow approximately linearly with the tree budget within the operating range. \saba{} therefore models the decoding cost as an affine function
\begin{equation}
C_t(B)=c_0+c_1B+c_2B\mathrm{prefix}_t,
\label{eq:saba-cost}
\end{equation}
where $c_0$ captures fixed per-round overhead, $c_1B$ the near-linear work from additional nodes, $c_2B\mathrm{prefix}_t$ the increasing attention cost of verifying a tree against a longer prefix.


\paragraph{Utility-based budget selection.} We fit the parameters of $G_t(B)$ and $C_t(B)$ using a small held-out validation set and keep them fixed during evaluation. At each decoding round, \saba{} evaluates $U_t(B)$ over budgets ranging from 32 to 512 in increments of 16 and selects $B_t=\arg\max_B U_t(B)$. Since the required state signals are already available from the drafter and previous verification rounds, \saba{} requires only lightweight scalar computation and adds approximately $0.1$\,ms per round, which is negligible compared with decoding latency.

\section{Experiments}

\begin{table*}[t]
  \centering
  \scriptsize
  \setlength{\tabcolsep}{2.0pt}
  \caption{\textbf{Main results with mean accepted tokens and wall-time speedup.} Sp. denotes TPS normalized by autoregressive decoding on the same model and dataset. Bold values mark the best MAT/Sp. within each model--dataset group.}
  \label{tab:main-mat-speedup}
  \resizebox{\textwidth}{!}{%
  \begin{tabular}{llcccccccccccccccccc}
  \toprule
  \multirow{2}{*}{Model} & \multirow{2}{*}{Method} & \multicolumn{2}{c}{GSM8K} & \multicolumn{2}{c}{Math500} & \multicolumn{2}{c}{AIME24} & \multicolumn{2}{c}{AIME25} & \multicolumn{2}{c}{HumanEval} & \multicolumn{2}{c}{MBPP} & \multicolumn{2}{c}{LCB} & \multicolumn{2}{c}{SWE-bench} & \multicolumn{2}{c}{Alpaca} \\
  \cmidrule(lr){3-4}\cmidrule(lr){5-6}\cmidrule(lr){7-8}\cmidrule(lr){9-10}\cmidrule(lr){11-12}\cmidrule(lr){13-14}\cmidrule(lr){15-16}\cmidrule(lr){17-18}\cmidrule(lr){19-20}
    & & MAT & Sp. & MAT & Sp. & MAT & Sp. & MAT & Sp. & MAT & Sp. & MAT & Sp. & MAT & Sp. & MAT & Sp. & MAT & Sp. \\
  \midrule
  \multirow{5}{*}{Qwen3-4B} & EAGLE-3 & 2.28 & 1.12 & 2.21 & 1.08 & 2.12 & 1.18 & 2.13 & 1.19 & 2.21 & 1.05 & 2.26 & 1.11 & 2.14 & 1.01 & 2.09 & 0.99 & 2.17 & 1.05 \\
    & OPT-Tree & 2.38 & 1.05 & 2.32 & 1.03 & 2.14 & 1.32 & 2.16 & 1.33 & 2.34 & 0.95 & 2.37 & 1.06 & 2.27 & 0.98 & 2.19 & 0.97 & 2.30 & 1.03 \\
    & \dflash{} & 5.87 & 4.11 & 7.07 & 5.09 & 6.08 & 4.36 & 6.76 & 4.91 & 6.00 & 4.30 & 5.40 & 3.84 & 6.13 & 4.47 & 3.29 & 2.30 & 2.85 & 2.23 \\
    & \ddtree{} & 8.10 & 5.16 & 9.37 & 5.98 & 7.39 & 5.26 & 7.86 & 5.54 & \textbf{8.54} & 5.36 & \textbf{7.82} & 5.09 & 8.45 & 5.16 & 5.04 & 3.08 & 4.45 & 2.96 \\
    & \graft{} & \textbf{8.18} & \textbf{5.58} & \textbf{9.68} & \textbf{6.36} & \textbf{7.99} & \textbf{5.34} & \textbf{8.45} & \textbf{5.72} & 8.41 & \textbf{5.66} & \textbf{7.82} & \textbf{5.21} & \textbf{8.46} & \textbf{5.75} & \textbf{5.21} & \textbf{3.41} & \textbf{4.50} & \textbf{3.10} \\
  \midrule
  \multirow{5}{*}{Qwen3-8B} & EAGLE-3 & 4.07 & 1.79 & 4.12 & 1.79 & 3.32 & 1.87 & 3.30 & 1.86 & 4.34 & 1.77 & 4.36 & 1.92 & 3.57 & 1.57 & 3.22 & 1.39 & 3.75 & 1.71 \\
    & OPT-Tree & 3.96 & 1.46 & 3.91 & 1.45 & 3.23 & 1.87 & 3.22 & 1.85 & 4.21 & 1.35 & 4.27 & 1.57 & 3.45 & 1.30 & 3.12 & 1.21 & 3.66 & 1.46 \\
    & \dflash{} & 5.82 & 3.90 & 7.02 & 4.84 & 6.15 & 4.18 & 6.62 & 4.51 & 6.43 & 4.46 & 5.49 & 3.76 & 6.08 & 4.30 & 3.28 & 2.16 & 2.83 & 2.02 \\
    & \ddtree{} & \textbf{8.18} & 4.33 & \textbf{9.38} & 4.93 & 7.25 & 4.99 & 7.80 & 5.35 & \textbf{9.13} & 4.57 & \textbf{7.93} & 4.26 & \textbf{8.55} & 4.41 & 5.04 & 2.56 & \textbf{4.46} & 2.38 \\
    & \graft{} & 8.06 & \textbf{5.28} & 9.36 & \textbf{6.17} & \textbf{8.04} & \textbf{5.45} & \textbf{8.63} & \textbf{5.85} & 8.96 & \textbf{5.87} & 7.83 & \textbf{5.25} & 8.32 & \textbf{5.33} & \textbf{5.08} & \textbf{3.13} & 4.26 & \textbf{2.80} \\
  \midrule
  \multirow{5}{*}{Qwen3-Coder-30B} & EAGLE-3 & 2.18 & 1.20 & 2.25 & 1.27 & 2.16 & 1.23 & 2.13 & 1.22 & 2.48 & 1.35 & 2.52 & 1.33 & 2.03 & 1.12 & 2.05 & 1.12 & 1.90 & 1.08 \\
    & OPT-Tree & 2.23 & 1.27 & 2.29 & 1.35 & 2.23 & 1.32 & 2.18 & 1.29 & 2.54 & 1.43 & 2.57 & 1.41 & 2.10 & 1.21 & 2.12 & 1.21 & 1.98 & 1.17 \\
    & \dflash{} & 4.72 & 3.16 & 4.96 & 3.45 & 4.50 & 3.21 & 4.58 & 3.17 & 7.39 & 4.90 & 7.02 & 4.50 & 5.41 & 3.91 & 3.08 & 2.07 & 2.09 & 1.61 \\
    & \ddtree{} & 6.83 & 4.01 & 7.18 & 4.37 & 5.70 & 3.82 & 6.02 & 3.93 & \textbf{10.06} & 5.70 & 9.38 & 5.11 & 7.67 & 4.53 & 4.78 & 2.81 & \textbf{3.22} & 2.11 \\
    & \graft{} & \textbf{7.24} & \textbf{4.28} & \textbf{7.45} & \textbf{4.74} & \textbf{6.61} & \textbf{4.33} & \textbf{6.87} & \textbf{4.42} & 10.01 & \textbf{6.01} & \textbf{9.83} & \textbf{5.56} & \textbf{7.93} & \textbf{4.91} & \textbf{5.04} & \textbf{3.03} & 2.96 & \textbf{2.13} \\
  \bottomrule
  \end{tabular}%
  }
\end{table*}

\subsection{Experimental Setup}

\paragraph{Tasks, models, and metrics.} We evaluate \graft{} on Qwen3-4B, Qwen3-8B, and Qwen3-Coder-30B-A3B-Instruct \cite{DBLP:journals/corr/abs-2505-09388} across nine benchmarks: GSM8K, Math500, AIME24, and AIME25 for mathematical reasoning \citep{DBLP:journals/corr/abs-2110-14168,DBLP:conf/nips/HendrycksBKABTS21,DBLP:journals/corr/abs-2503-21380}; HumanEval, MBPP, and LiveCodeBench for code generation \citep{DBLP:journals/corr/abs-2107-03374,DBLP:journals/corr/abs-2108-07732,DBLP:conf/iclr/JainHGLYZWSSS25}; SWE-bench for repository-level software engineering \citep{DBLP:conf/iclr/JimenezYWYPPN24}; and Alpaca for instruction following \cite{taori2023alpaca}. Unless otherwise specified, all experiments use a A100 80GB GPU with batch size 1. We report mean accepted tokens (MAT), tokens per second (TPS), wall-time speedup over autoregressive decoding, and mismatch reduction.

\paragraph{Baselines.} We compare against five baselines. \textbf{(1) Autoregressive Decoding}: the target-model throughput baseline. \textbf{(2) EAGLE-3}\cite{li2026eagle}: a strong autoregressive drafter with draft-tree verification. \textbf{(3) OPT-Tree}\cite{DBLP:journals/tacl/WangSLXYDWZ25}: an adaptive tree-construction method built on the EAGLE-3 drafter. \textbf{(4) \dflash{}}\cite{DBLP:journals/corr/abs-2602-06036}: a DLM drafter that predicts future positions in one forward pass. \textbf{(5) \ddtree{}}\cite{DBLP:journals/corr/abs-2604-12989}: the DLM-based draft-tree baseline that assembles a fixed-budget tree from \dflash{} token distributions.

\paragraph{Implementation details.} For EAGLE-3, OPT-Tree, and \ddtree{}, we follow the settings in their official repositories. For \tdes{}, we empirically set $\beta=0.2$ by default. \saba{} fits the parameters of $G_t(B)$ and $C_t(B)$ on held-out validation prompts through same-prefix budget replay, and the fitted parameters are fixed during evaluation.

\paragraph{TDES training.} We train \tdes{} on target-model traces from 3,046 held-out prompts drawn from the seven non-AIME benchmarks. AIME24 and AIME25 are used only for evaluation due to their small sample sizes. The scorer is trained for 8 epochs with AdamW, learning rate $2\times10^{-3}$, weight decay $10^{-4}$, and loss weights $\lambda_{\mathrm{KL}}=1$, $\lambda_{\mathrm{pair}}=1$, and $\lambda_{\mathrm{BCE}}=0.25$.

\subsection{Main Results}

We make three observations from Table~\ref{tab:main-mat-speedup}. \textbf{(i)} \graft{} achieves the highest wall-time speedup in all 27 model--dataset groups. The best speedups on Qwen3-4B, Qwen3-8B, and Qwen3-Coder-30B are $6.36\times$, $6.17\times$, and $6.01\times$, respectively. \textbf{(ii)} Higher speedup does not always require higher MAT. In several groups, \graft{} has slightly lower MAT than \ddtree{} because \graft{} optimizes throughput rather than accepted length alone. Some additional tree nodes can increase MAT but still reduce TPS due to their verification cost. \textbf{(iii)} The gains hold across task categories. \graft{} improves speedup on mathematical reasoning, code generation, software engineering, and instruction following. The AIME results provide an additional generalization check since AIME24 and AIME25 are excluded from \tdes{} training.

\subsection{Ablation Studies}

We seek to answer the following questions in ablation studies. \textit{(1) How much does each component contribute when the other is held fixed or removed?} \textit{(2) Does \tdes{} generalize across task categories rather than learning dataset-specific shortcuts?} \textit{(3) How closely does \saba{} track the oracle budget?} \textit{(4) How sensitive is the residual edge correction to the choice of $\beta$?} Unless otherwise specified, all ablation studies are conducted under the Qwen3-8B setting, using Qwen3-8B as the target model and its corresponding \dflash{} drafter.

\paragraph{Component Analysis.} Table~\ref{tab:component-analysis} isolates the two decisions made by \graft{}, namely edge scoring and budget allocation. With the same fixed budget as \ddtree{}, +\tdes{} improves MAT from $7.53$ to $8.00$ and TPS by $5.7\%$, confirming that target-distilled edge scores improve tree quality without changing tree size. With the \ddtree{} edge score, +\saba{} reduces the average tree size from $110.2$ to $49.1$ nodes and still improves TPS by $12.8\%$, showing that many fixed-budget nodes have limited throughput utility. The full \graft{} combines \tdes{}'s improved tree quality with \saba{}'s budget efficiency. Compared with +\saba{}, it restores MAT from $7.18$ to $7.62$ at nearly the same budget and reaches the highest TPS of $263.5$.

\begin{table}[t]
  \centering
  \caption{\textbf{Component analysis on Qwen3-8B.} Results are averaged over the nine evaluation datasets. +\tdes{} replaces the \ddtree{} edge score with \tdes{} while keeping the fixed \ddtree{} budget. +\saba{} keeps the \ddtree{} edge score and selects the tree budget with \saba{}. \graft{} enables both components. Avg. Nodes denotes the average number of draft-tree nodes per round.}
  \label{tab:component-analysis}
  \resizebox{\columnwidth}{!}{%
  \begin{tabular}{lcccc}
    \toprule
    Method & Avg. Nodes & MAT & TPS & $\Delta$TPS vs. \ddtree{} \\
    \midrule
    \ddtree{} & 110.2 & 7.53 & 220.6 & -- \\
    +\tdes{} & 110.2 & \textbf{8.00} & 233.3 & +5.7\% \\
    +\saba{} & 49.1 & 7.18 & 248.8 & +12.8\% \\
    \graft{} & 48.9 & 7.62 & \textbf{263.5} & \textbf{+19.4\%} \\
    \bottomrule
  \end{tabular}%
  }
\end{table}

\begin{figure}[t]
  \centering
  \includegraphics[width=\columnwidth]{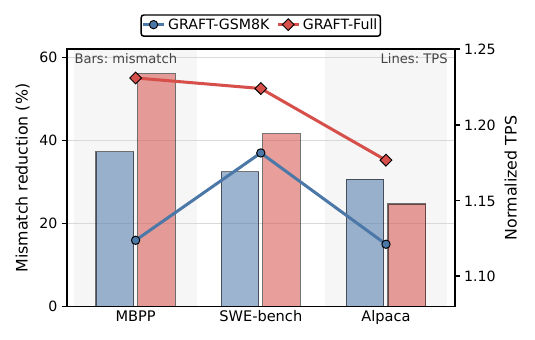}
  \caption{\textbf{Cross-task generalization of \tdes{}.} GRAFT-GSM8K uses a scorer trained only on GSM8K traces. GRAFT-Full uses the scorer from the main experiments. Both variants enable \saba{}. Bars report parent--child mismatch reduction, while lines report TPS normalized by \ddtree{}.}
  \label{fig:tdes-generalization}
\end{figure}

\paragraph{TDES Generalization.} The AIME24 and AIME25 results in Table~\ref{tab:main-mat-speedup} show that \graft{} remains effective on mathematical reasoning datasets excluded from \tdes{} training. This evidence is still limited, as these datasets share the same broad task category as part of the training data. We therefore train the edge scorer only on GSM8K traces and evaluate the resulting \graft{} variant on MBPP, SWE-bench, and Alpaca, which cover code generation, software engineering, and instruction following. As shown in Figure~\ref{fig:tdes-generalization}, GRAFT-GSM8K consistently reduces parent--child mismatch and improves normalized TPS on all three tasks. This suggests that \tdes{} captures reusable parent--child compatibility rather than a purely dataset-specific shortcut.

\begin{figure}[t]
  \centering
  \includegraphics[width=\columnwidth]{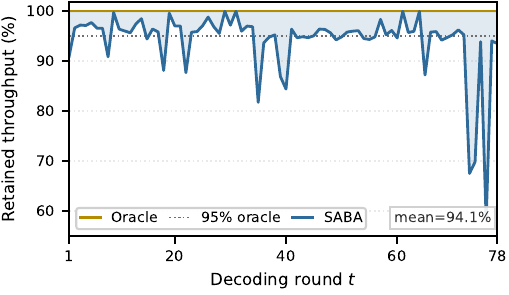}
  \caption{\textbf{Same-prefix oracle tracking of \saba{}.} A randomly selected GSM8K prompt is decoded once by \graft{} to form the reference trajectory. Each round is replayed with budgets $\{32,48,64,80,96,128\}$ while keeping the prefix, KV cache, and \dflash{} outputs unchanged. The curve reports retained oracle throughput $\mathrm{TPS}_t(B_t)/\mathrm{TPS}_t(B_t^\star)$.}
  \label{fig:saba-oracle-tracking}
\end{figure}

\paragraph{SABA Oracle-Tracking Analysis.} \saba{} should select a budget that maximizes per-round throughput. To measure how closely \saba{} approaches the oracle budget, we randomly select one GSM8K prompt, run \graft{} once, and fix the resulting reference trajectory. At each round, we replay every budget in $\mathcal{B}=\{32,48,64,80,96,128\}$ using the same prefix, KV cache, and \dflash{} outputs. For each budget, we compute the per-round throughput as $\mathrm{TPS}_t(B)=a_t(B)/\ell_t(B)$, where $a_t(B)$ is the accepted-token count and $\ell_t(B)$ is the measured round latency. The same-prefix oracle is $B_t^\star=\arg\max_{B\in\mathcal{B}}\mathrm{TPS}_t(B)$. As shown in Figure~\ref{fig:saba-oracle-tracking}, \saba{} retains $94.1\%$ of oracle throughput on average and reaches at least $95\%$ of oracle throughput in $64.1\%$ of rounds.

\paragraph{Hyperparameter Sensitivity.} Since \tdes{} adds a residual compatibility score on top of the \dflash{} marginal, $\beta$ controls the trade-off between parent--child compatibility and marginal path coverage. We evaluate $\beta$ in Table~\ref{tab:beta-sensitivity} using full \graft{} on GSM8K. Setting $\beta=0$ disables the \tdes{} residual while keeping \saba{} active. The settings $\beta\in\{0.05,0.1,0.2\}$ all improve MAT and TPS over $\beta=0$. The best setting is $\beta=0.2$, which reaches $8.061$ MAT and $276.17$ TPS with a $54.71\%$ mismatch reduction. When $\beta$ is increased to $0.4$, mismatch reduction rises to $79.32\%$, but MAT and TPS fall below $\beta=0$.  This drop shows that mismatch reduction alone is not the optimization target. Under a finite tree budget, an overly strong residual can over-prioritize edges with high compatibility scores and discard high-marginal paths that would otherwise cover the target sequence. We therefore use $\beta=0.2$ as the default setting.

\begin{table}[t]
  \centering
  \caption{\textbf{Sensitivity to the residual scale $\beta$ on GSM8K.} All rows use the same \saba{} controller. $\beta=0$ corresponds to \ddtree{} edge scoring with \saba{}. Mismatch Red. is computed relative to $\beta=0$.}
  \label{tab:beta-sensitivity}
  \begin{tabular}{lccc}
    \toprule
    $\beta$ & MAT & TPS & Mismatch Red. (\%) \\
    \midrule
    0 & 7.635 & 260.19 & 0.00 \\
    0.05 & 7.850 & 262.82 & 22.14 \\
    0.1 & 7.968 & 267.01 & 33.21 \\
    0.2 & \textbf{8.061} & \textbf{276.17} & 54.71 \\
    0.4 & 7.351 & 247.60 & 79.32 \\
    \bottomrule
  \end{tabular}
\end{table}

\section{Related Work}

\paragraph{Speculative decoding.} Speculative decoding accelerates autoregressive generation by predicting candidate tokens and verifying them in parallel \citep{DBLP:journals/corr/abs-2302-01318,DBLP:conf/icml/LeviathanKM23}. Existing approaches can be broadly divided into model-free and model-based drafters. Model-free methods avoid training an auxiliary drafter and instead construct proposals from reusable text patterns or retrieval sources \cite{DBLP:conf/icml/FuBS024,DBLP:conf/iclr/LiuLLLZHS25,DBLP:conf/naacl/0012ZCL024}. For example, SuffixDecoding \cite{oliaro2024suffixdecoding} builds suffix trees over prompts and previous outputs to generate candidate sequences. Model-based methods learn an additional proposal mechanism and often achieve stronger draft accuracy \cite{DBLP:conf/icml/LiW0024,DBLP:conf/emnlp/LiW0024}. Medusa \cite{DBLP:conf/icml/CaiLGPLCD24} attaches lightweight decoding heads to the target model, while EAGLE-3 \citep{li2026eagle} draft from target-model features. More recently, some works focus on DLM-based drafter that predicts multiple future positions in one pass \cite{DBLP:journals/corr/abs-2511-00606}. \dflash{} \citep{DBLP:journals/corr/abs-2602-06036} shows the promise of this approach, achieving state-of-the-art speculative decoding performance.


\paragraph{Tree-based speculative decoding.} Existing tree-based methods mainly differ in how they construct the draft tree. For autoregressive drafters, tree-based methods adapt the tree shape or budget under parent-conditioned expansion using confidence and cost signals \citep{DBLP:conf/asplos/MiaoOZCWZWZYSSC24,DBLP:conf/icml/CaiLGPLCD24,DBLP:conf/emnlp/LiW0024,li2026eagle,DBLP:journals/tacl/WangSLXYDWZ25,DBLP:journals/corr/abs-2604-09603,shen2026specbranch}. For instance, TALON \cite{DBLP:journals/corr/abs-2601-07353} allocates a fixed node budget between depth and width according to drafting confidence, while CAST \cite{DBLP:journals/corr/abs-2510-26577} refines the tree structure by modeling inference costs such as GPU devices and batch sizes. For DLM drafters, \ddtree{} constructs a fixed-budget draft tree from \dflash{} per-position distribution with heap expansion \citep{DBLP:journals/corr/abs-2604-12989}.

\section{Conclusion}

We propose \graft{}, a draft-tree construction framework for \dlm{}-based speculative decoding. \graft{} addresses two limitations of \ddtree{}'s parent-agnostic, fixed-budget tree assembly. Target-distilled edge scoring improves which edges enter the tree, while state-aware budget allocation determines how many nodes should be verified in each round. Experiments show that \graft{} achieves $2.13\times$--$6.36\times$ speedup over autoregressive decoding. Although \graft{} improves the quality and budget efficiency of \dlm{} draft trees, it still executes drafting, tree construction, and target verification sequentially at each round. Reducing the resulting pipeline bubbles through concurrent speculative decoding remains an interesting direction for future work.

\bibliography{aaai2027}


\end{document}